\documentclass[lettersize,journal]{IEEEtran}
\usepackage{amsmath,amsfonts}
\usepackage{algorithmic}
\usepackage{algorithm}
\usepackage{array}
\usepackage[caption=false,font=normalsize,labelfont=sf,textfont=sf]{subfig}
\usepackage{textcomp}
\usepackage{stfloats}
\usepackage{url}
\usepackage{verbatim}
\usepackage{graphicx}
\usepackage{cite}
\usepackage{color}
\usepackage[table]{xcolor}

\usepackage{xspace}
\usepackage{expl3}
\usepackage{multirow}
\usepackage{float} 
\usepackage{caption}
\usepackage{listings}
\usepackage{hyperref}
\hypersetup
{
    colorlinks = true,
    linkcolor = red,
    citecolor = green,
}
\usepackage{bbding}
\usepackage{booktabs}
\usepackage{makecell}

\begin{document}

\title{Semantic-Constraint Matching Transformer for Weakly Supervised Object Localization}

\author{Yiwen Cao, Yukun Su, Wenjun Wang, Yanxia Liu and Qingyao Wu
\thanks{Y. Cao, Y. Su, W. Wang, Y. Liu and Q. Wu are with the School of Software Engineering, South China University of Technology, Guangzhou 510640, China, and also with the Key Laboratory of Big Data and Intelligent Robot, Ministry of Education, Guangzhou, China, E-mail: yiwen.cao@outlook.com}
\thanks{Y. Su is also with the School of Computer Science and Engineering, Nanyang Technological University, Singapore.}
\thanks{Q. Wu is also with the Pazhou Lab, Guangzhou, China.}
\thanks{Y. Liu and Q. Wu are the corresponding authors. E-mail: qyw@scut.edu.cn.}
}

\markboth{Journal of \LaTeX\ Class Files,~Vol.~14, No.~8, January~2023}%
{Shell \MakeLowercase{\textit{et al.}}: A Sample Article Using IEEEtran.cls for IEEE Journals}


\maketitle

\begin{abstract}
Weakly supervised object localization (WSOL) strives to learn to localize objects with only image-level supervision. Due to the local receptive fields generated by convolution operations, previous CNN-based methods suffer from partial activation issues, concentrating on the object's discriminative part instead of the entire entity scope. Benefiting from the capability of the self-attention mechanism to acquire long-range feature dependencies, Vision Transformer has been recently applied to alleviate the local activation drawbacks. However, since the transformer lacks the inductive localization bias that are inherent in CNNs, it may cause a divergent activation problem resulting in an uncertain distinction between foreground and background. 
In this work, we proposed a novel 
\textbf{S}emantic-\textbf{C}onstraint \textbf{M}atching \textbf{N}etwork (SCMN) via transformer to converge on the divergent activation. 
Specifically, we first propose a local patch shuffle strategy to construct the image pairs, disrupting local patches while guaranteeing global consistency. The paired images that contain the common object in spatial are then fed into the Siamese network encoder. We further design a semantic-constraint matching module, which aims to mine the co-object part by matching the coarse class activation maps (CAMs) extracted from the pair images, thus
implicitly guiding and calibrating the transformer network to alleviate the divergent activation. 
Extensive experimental results conducted on two challenging benchmarks including CUB-200-2011 and ILSVRC datasets show that our method can achieve the new state-of-the-art performance and outperform the previous method by a large margin.
\end{abstract}


\section{Introduction}
\label{sec1}

\begin{figure}[!t]
  \centering
  \includegraphics[width=\columnwidth]{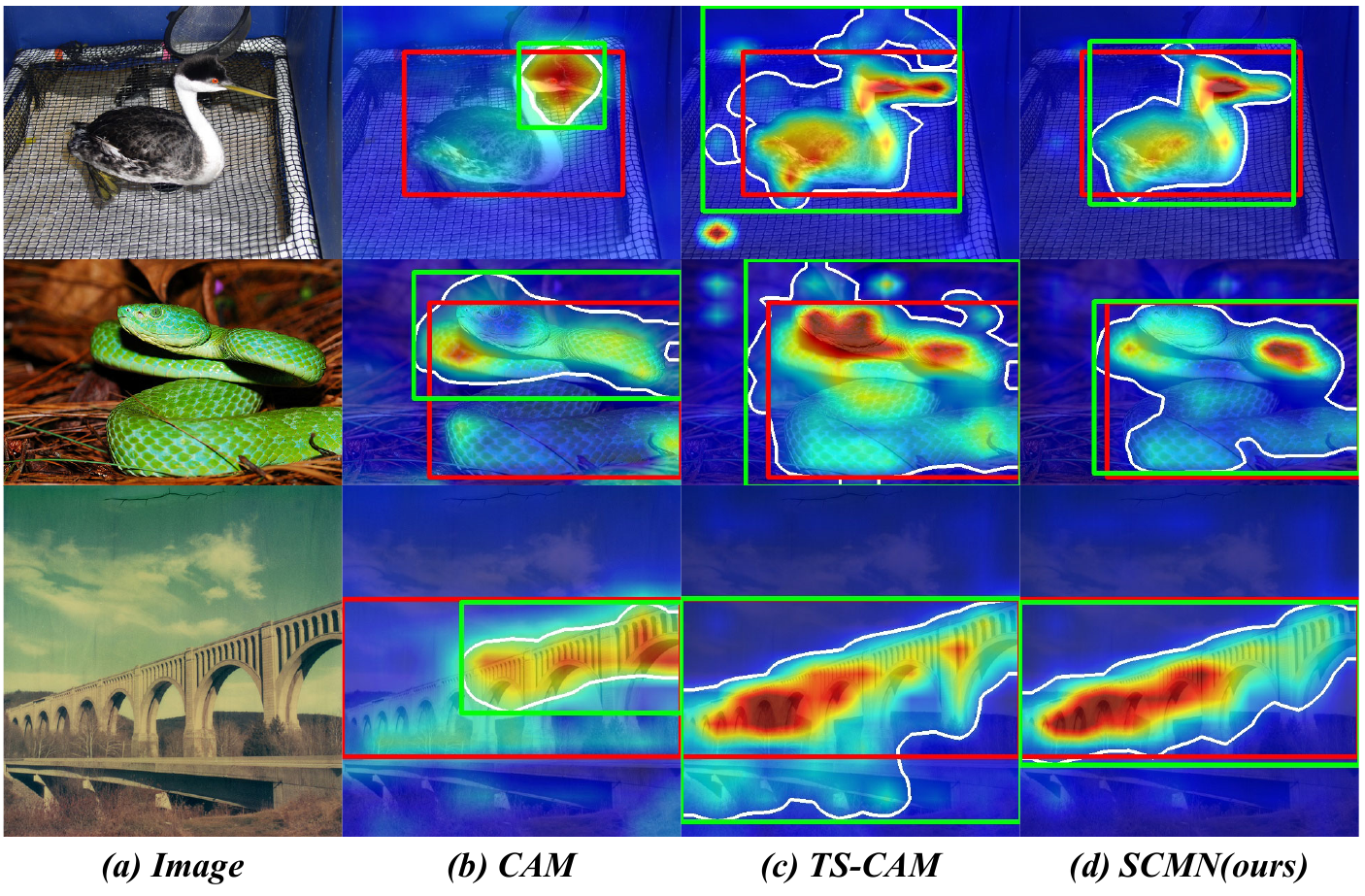}
  \caption{\textbf{Visualization of semantic activation maps and comparison of localization results}, where {\color{green}green} and {\color{red}red} boxes indicate the network prediction and ground-truth, respectively. The white line shows the contour of the foreground. (a) Original image. (b) CNN-based methods suffer from partial activation problem, anchoring local regions of objects. (c) Transformer-based method is plagued by divergent activation issue resulting in an uncertain distinction between foreground and background. (d) SCMN activates the entire entity while suppressing background divergence activation.}
  \label{Fig.main_1}
\end{figure}

\IEEEPARstart{O}{bject} detection and localization are the core tasks in computer vision, which have achieved remarkable progress with fully supervised deep learning networks.
Weakly supervised learning (WSL) method utilizes coarse annotations or weak labels for model learning~\cite{zhou2018brief}, which has mature developments in areas such as Temporal Action Localization~\cite{zhang2019glnet, su2020transferable, zhai2021action} and Semantic Segmentation~\cite{zhang2019decoupled, zhou2020sal, su2021context}.
Weakly supervised object localization (WSOL) benefiting significantly from WSL, uses off-the-shelf and simple image-level classification labels to replace detailed location annotations (e.g., bounding boxes). 
Accordingly, WSOL has broad application opportunities in various scenarios, such as image retrieval, medical lesion localization, etc.


The seminal work (Zhou $\emph{et~al.}$~\cite{zhou2016learning}) converts conventional classification networks 
by replacing the top layer of the network with a global average pooling (GAP)~\cite{lin2013network} layer. As a result, it can assemble the output feature space of the last convolutional layer to generate a class activation map (CAM) for localization purposes. However, as the network seeks to attain higher classification performance, it tends to focus on the significantly discriminative parts of the objects. Also, the CNN-based approach can only access short-range feature dependencies due to the limited receptive field. In that case, its corresponding class activation map activates more local regions of the entity, decreasing the performance of object localization, as shown in Figure \ref{Fig.main_1}(b).


Recently, Transformer~\cite{vaswani2017attention}, the cornerstone work in natural language processing (NLP), has been involved in various computer vision tasks with superior accomplishments.
Vision transformer segments the image into a patch sequence as input and gets excellent visual embedding output by cascading Multilayer Perceptron (MLP) structure. Furthermore, Transformer can acquire long-range feature dependencies thanks to its intrinsic self-attention mechanism, which fundamentally avoids the local activation problem of CNN-based structures in WSOL. TS-CAM~\cite{gao2021ts} first directly applies the transformer structure to WSOL, which couples the patch token features with the semantic-agnostic attention map for object localization with promising results. However, Transformer's global viewing capability allows it to reference both the background and the foreground. Also, many categories of objects and backgrounds are strongly correlated, such as "duck" and "river", "car" and "road " etc.
As shown in Figure \ref{Fig.main_1}(c), the lack of inductive localization bias inherent to CNN leads to the problem of divergent activation, implying that simultaneous activation of foreground and background lead to mislabeling the candidate bounding box. Therefore, suppressing or converging the divergent activation is a new challenge the transformer structure brings.

\begin{figure}[!t]
    \centering
    \includegraphics[width=\columnwidth]{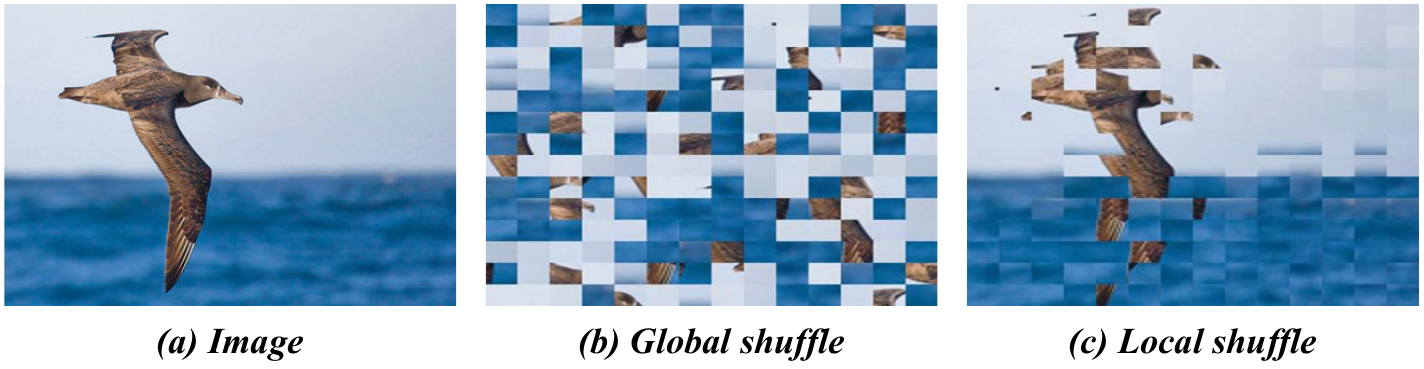}
    \caption{\textbf{Comparison of global and local shuffle mechanisms.} (a) Original image. (b) Global shuffled image, which eliminates low-level semantic information. (c) Local shuffled image, which maintains the object spatial and semantic consistency.}
    \label{Fig.main_2}
\end{figure}

To this end, we propose a novel Semantic-Constraint Matching Network (SCMN) built upon the Transformer to converge the divergent activation via end-to-end weakly supervised training. 
Our central concept is to discover the co-activation object part by matching semantic activation maps between the primal image and the image after patches-shuffling, thus implicitly guiding and calibrating the transformer network through loss functions to alleviate the divergent activation.
With regards to this, we first proposed a local patch shuffle strategy to construct the primal-shuffled image pairs as shown in \ref{Fig.main_2}. Since the input to the Transformer network is a sequence of patches, scrambling the sequence can result in various highlighting effects in the semantic activation map. 
Subsequently, we design a semantic-constraint matching module to acquire a flow matrix that matches the co-activation part of the semantic activation maps generated from the primal-shuffled image pair.
The network focuses on the entire foreground area when learning general semantic knowledge from co-activated regions. Utilizing features altered by the flow matrix to construct the loss function, we indirectly refine the Transformer's output embedding to calibrate the network for restricting divergent activation, as shown in Figure \ref{Fig.main_1}(d).
To validate the effectiveness of the proposed SCMN, extensive experiments are conducted on challenging WSOL benchmarks. To show its robustness, we further apply it to address weakly supervised salient object detection. Experimental results show the superiority of our proposed method, and we can outperform the previous method by a large margin.

The main contributions of our paper can be summarized as follows:
\begin{itemize}
\item We take an early attempt to experimentally investigate the transformer network architecture in WSOL field, which is not well-explored yet. 
\item We propose a novel Semantic-Constraint Matching transformer Network with a tailored local patch shuffle strategy and semantic-constraint matching module to restrain divergent additional background activations. 
\item Experimental evaluations on two challenging benchmarks including CUB-200-2011 and ILSVRC2012 dataset show the effectiveness of our proposed approach and achieve new state-of-the-art performance.
\end{itemize}

\section{Related Work}

\subsection{Weakly supervised object localization via CNN}~\label{sec.2.1}
Weakly supervised object localization (WSOL) is intended to accomplish localization with bounding boxes using category image-level labels. Many CNN-based methods have continually improved and revised to provide competitive results. Zhou $\emph{et~al.}$~\cite{zhou2016learning} first discovered that the feature representation of the classification network and replaced the top fully connected layer with a global average pooling layer to gain class activation maps (CAMs), representing each region's contribution to the classification result. 
However, it fails to locate the whole entity under local activation problem. 
A typical technique to improve the localization capacity of CAM is erasing. HaS~\cite{singh2017hide} delivers a regularization method for random hidden image patches similar to DropBlock~\cite{ghiasi2018dropblock}, which strives to have the network seek other non-remarkable components of the object. 
ACoL~\cite{zhang2018adversarial} presents a double-classifiers network while one uncovers discriminative parts and the other erases discovered areas to force the network to find complementary object elements. ADL~\cite{choe2019attention} subsequently offers an attention-guided dropout pipeline that withdraws the pixels of the feature layer that are of deep concern. 
In addition to erasure, other practices have also reached sounder marks in WSOL. DANet~\cite{xue2019danet} performs divergent activation of hierarchical features and integrates multi-layer features to obtain full object activation. 
SPG~\cite{zhang2018self} introduced the constraint of pixel-level correlations by an attention-based mask. I$^2$C~\cite{zhang2020inter} and SLT-Net~\cite{guo2021strengthen} both introduce a multi-input co-activation approach, and the latter separates WSOL into two sub-tasks. Tao $\emph{et~al.}$~\cite{tao2018exploiting} presented a transfer learning method to extract prior knowledge from a collection of web images.
\begin{figure*}[h]
  \centering
  \includegraphics[width=\textwidth]{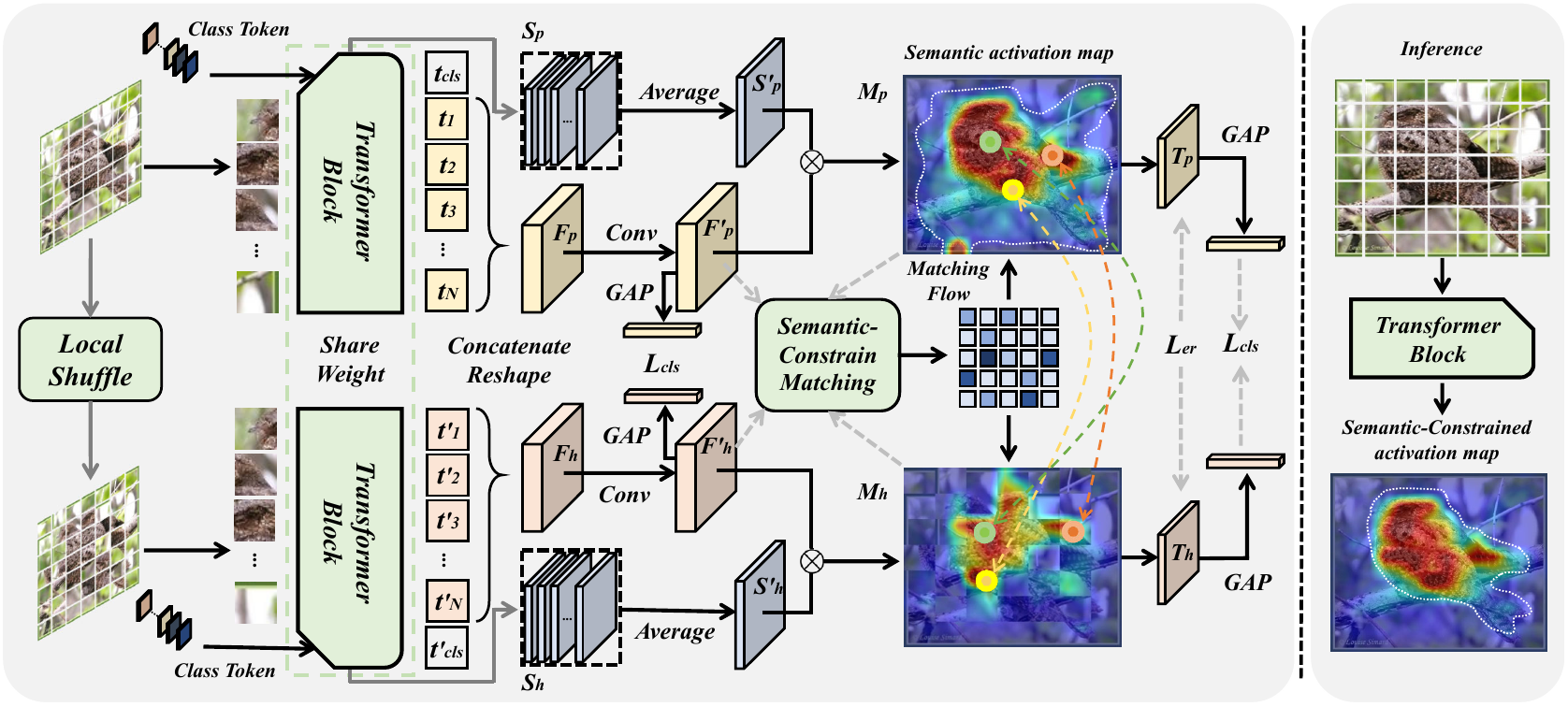}
  \caption{\textbf{Pipeline of the proposed Semantic-Constraint Matching Transformer Network.} (Left) Given an input image $I$, we use a local patch shuffle strategy to construct primal-shuffled pair. The paired images that contain the common object in spatial are then fed into the shared-weight transformer encoder. By multiplying the inner-guided attention maps $S'_p$, $S'_h$ and analogous class activation maps $F'_p$, $F'_h$, we implemented semantic activation maps $M_p$ with divergent activation problem (shown as the white dashed line) and $M_h$. Next, we use a semantic-constraint matching module, which takes $M_p$, $M_h$ as references and $F'_p$, $F'_h$ as input to compute a matching flow matrix for obtaining refined feature $T_p$, $T_h$. Finally, we calibrate the network constraint divergent activation by optimizing $\mathcal{L}_{cls}$ and $\mathcal{L}_{er}$. (Right) We directly apply calibrated Transformer network during inference time to obtain a flawless semantic-constrained activation map.}
  \label{Fig.main_3}
\end{figure*}
\subsection{Weakly supervised object localization via Transformer}~\label{sec.2.2}
Transformer~\cite{vaswani2017attention,su2023unified} is the preeminent sequence-to-sequence model in natural language processing and has been applied as a general-purpose backbone in computer vision rapidly. 
TS-CAM~\cite{gao2021ts} as a guiding work first directly applied Transformer to WSOL. Similar to CAM, it treats the patch token as the feature representation of the output, completes the classification prediction through a GAP layer, and then combines it with the semantic-agnostic attention map in Transformer to complete the localization. Appreciating Transformer's ability of acquiring long-range feature dependencies, TS-CAM gets rid of the problem of local activation. 
However, as we mentioned in Sec.\ref{sec1}, the inductive localization bias embedded in CNN stems from the local receptive field and translation invariance due to its unique convolution and max-pooling operations. Due to the lack of this property, TS-CAM with straightforward application of Transformer inevitably suffers from the crisis of divergent activation, i.e., false activation of the background.
Compared with existing pipeline, our SCMN uses semantic matching to alleviate divergent activation while preserving long-range dependencies based on transformer architecture that is beneficial for WSOL.



\subsection{Image matching}~\label{sec2.2}
Image matching or semantic correspondence target to establish correlation among critical points or regions between images, which has long been one of the fundamental tasks of computer vision. 
The image matching mechanism has vast applications, such as Simultaneous Mapping (SLAM)~\cite{endres2012evaluation}, semantic segmentation~\cite{rubinstein2013unsupervised,taniai2016joint,huo2022dual, su2022self}, pose transformation~\cite{zhou2021cocosnet, su2022general} and image editing~\cite{szeliski2007image,lin2011smoothly}. 
Previously, standard pipelines operate geometric models~\cite{fischler1981random,lamdan1988object} with local hand-crafted descriptors like 
SIFT~\cite{lowe2004distinctive}. Recent approaches to deep learning frequently focus on leveraging convolutional neural networks to learn better image features and local descriptors~\cite{detone2018superpoint,sarlin2020superglue}. 
In recent years, new advances in graph matching have been made with the resurgence of optimal transport~\cite{villani2009optimal}. Optimal transport is a traditional generalized linear assignment problem that an uncomplicated and efficient Sinkhorn algorithm~\cite{sinkhorn1967concerning,peyre2019computational} can approximately solve. 
Liu $\emph{et~al.}$~\cite{liu2020semantic} apply a staircase function on class activation map for optimal transport. DeepEMD~\cite{zhang2020deepemd} adopts the Earth Mover’s Distance (EMD) for few-shot image classification. DeepACG~\cite{zhang2021deepacg} utilize the  Gromov-Wasserstein(GW) distance to build dense correlation volumes within the image group for co-saliency detection. In this paper, we design a semantic-constraint matching module that matches the output primal-shuffled features guided by semantic activation map pairs to find the object part and thus quell divergent activation.

\section{Method}
This section thoroughly explained the roles and combinations of the SCMN components. Specifically, we first give an overview of the entire architecture and then introduce the local patch shuffle strategy and semantic-constraint matching module in detail. Ultimately, we outline the network training loss function and demonstrate how to make predictions about the final localization outcomes using the suggested framework during inference.

\subsection{Overall architecture}
As stated in Sec.\ref{sec1}, the global feature dependence of the Transformer structure enhances the correlation between foreground and background, strengthening the divergent activation problem. Our proposed SCMN implicitly calibrates the Transformer to constrain the divergent activation by matching the co-activation regions of the primal-shuffled image pairs.
The entire framework of our Semantic-Constraint Matching Transformer Network (SCMN) is shown in Figure~\ref{Fig.main_3}. 

Formally, given an image $I$ of $H \times W$ resolution, we split it into $N = h \times w$  square patches, where the edge length of the patch is $P=H/h=W/w$.
The divided patches are laterally flattened and linearly projected to compose a patch token sequence with an extra learnable class token $X_{cls} \in \mathbb{R}^{1\times D}$, where $D$ represents the patch dimension. Next, we adopt a local patch shuffle strategy to build primal-shuffled image pair.
The paired patch sequences 
are then fed into the Siamese Transformer encoder with shared weight.

Denoted $(\{t_n \in \mathbb{R}^{1\times D}, n=1,2,...,N\}, t_{cls})$ 
and $(\{t'_n \in \mathbb{R}^{1\times D}, n=1,2,...,N\}, t'_{cls})$
is the primal-shuffled output token embeddings of Transformer Block with $L$ layers. Inspired by TS-CAM~\cite{gao2021ts}, 
we 
re-allocate the semantic to patch token. Specifically, 
we concatenate the corresponding $N$ patch token ${t_n, t'_n}$ and reshape them individually to generate feature maps ${F_p,F_h \in \mathbb{R}^{D\times h\times w}}$. 
Similar to CAM~\cite{zhou2016learning}, after a $3\times3$ convolution layer, we obtained analogous class activation feature maps $F'_p, F'_h\in\mathbb{R}^{c\times h\times w}$, where $c$ denotes the number of categories.

Meanwhile, to generate reliable localization, 
we utilize the deep statistical properties of the class tokens for the guidance of patch tokens to generate an inner-guided attention map. For simplicity, we only detail the subsequent processing of the primal image, while shuffled image will undergo the same pipeline. Explicitly, at $l$-th transformer layer, the self-attention weight matrix ${W^l_p}\in\mathbb{R}^{(N+1)\times(N+1)}$ is averaged over multiple heads. Defining ${S^l_p}\in\mathbb{R}^{h\times w}$ as the attention weight corresponding to the class tokens in $W^l_p$, we average it over all intervening layers to acquire the inner-guided attention map $S'_p$.
\begin{equation}\label{eq1}
    S'_p = \frac{1}{L}{\textstyle \sum_{l=1}^{L} S^l_p} 
\end{equation}

Subsequently, we obtain $\hat{M}_p\in\mathbb{R}^{c\times h\times w}$ by coupling the analogues class activation map $F'_p$ and inner-guided attention map $S'_p$ using element-wise multiplication. A divergent primal semantic activation map $M_p\in\mathbb{R}^{h\times w}$ is obtained by extracting the $y$-th dimension of $\hat{M}_p$, where $y$ is the category label. Likewise, the shuffled input follow the same pipeline to obtain $M_h$. Finally, we use a semantic-constraint matching module, which takes $M_p,M_h$ as references and matches $F'_p,F'_h$ to calculate a matching flow matrix. The matrix contains the implicated knowledge of the co-object part in the primal-shuffled pair, thus guiding Transformer to curb the divergent activation by optimizing loss functions. More details are described in the subsequent chapters.

\subsection{Local patch shuffle strategy}~\label{sec3.2}

\begin{algorithm}[!t]
\caption{Local patch shuffle strategy}
\label{algrithm1}
\begin{algorithmic}[1]
\renewcommand{\algorithmicrequire}{ \textbf{Input:}} 
\renewcommand{\algorithmicensure}{ \textbf{Output:}} 
\REQUIRE 
primal image $I$, patch height/width $P$, image height $H$, image width $W$, probability $\eta$\\
Initialize $I_h = \emptyset$, $h=H/P$, $w=W/P$
\ENSURE 
shuffled image $I_h$
\STATE Split I into $h\times w$ patches;
\STATE $n=(h\times w)/4$
\FOR{$i=0$ to $n-1$}
\STATE $I_h^i = (I_{4i},I_{4i+1},I_{4i+2},I_{4i+3})$
\STATE $\eta_s=random(0,1)$
\IF{$\eta_s<\eta$}
\STATE ${I'}_h^i$ = random.shuffle($I_h^i$);
\STATE $I_h$.append(${I'}_h^i$);
\ELSE
\STATE $I_h$.append($I_h^i$);
\ENDIF
\ENDFOR
\STATE $I_h$.reshape($H,W$);
\STATE return $I_h$
\end{algorithmic}
\end{algorithm}

To establish primal-shuffled image pairs, we present a straightforward yet expedient affine transformation method for the vision transformer architecture, known as local patch shuffle strategy. As described in Sec.\ref{sec.2.2}, the transformer architecture emerged in NLP to deal with sequence-to-sequence problems. Considering that the position and order of words in a sentence are crucial, a word with different positions in a sentence may bring diametrically opposite meanings. Therefore, position encoding is introduced in Transformer to gain location information, defined in vision transformer as a set of learnable variables. So the sequence order also affects the outcome of network learning.

We are inspired by CNN's translation invariance, which means that the network can still recognize the interested target in image classification even if its appearance has been modified partially. Combined with the patch division in vision transformer, we suggest locally shuffling the image patches that implicitly enforce the network to learn translation invariance. Concretely, we perform a probabilistic shuffle of the contents of each quadrille image patch (a block containing four patches). The apiece image is split into $N/4$ quadrille blocks, each measuring $2P\times 2P$. We locally shuffle the four patches within a block at random without altering the order of the blocks as a whole. Hence it is possible to modify the patch order while maintaining the integrity of the global semantic information. Detailed solution is presented in Algorithm \ref{algrithm1}.

As shown in Figure \ref{Fig.main_2}, global shuffle corrupts entire image contents, significantly reducing network capability. In contrast, the local shuffle strategy protects the object's general semantic information and improves the robustness of the network. We expect the network to learn invariance implicitly and can intermittently focus more on object elements. The semantic-constraint matching module further processes the newly created semantic shuffle pairs in order to focus the network's attention on the foreground entirely.

\subsection{Semantic-constraint matching module}~\label{sec3.3}

\begin{figure}[!t]
    \centering
    \includegraphics[width=\columnwidth]{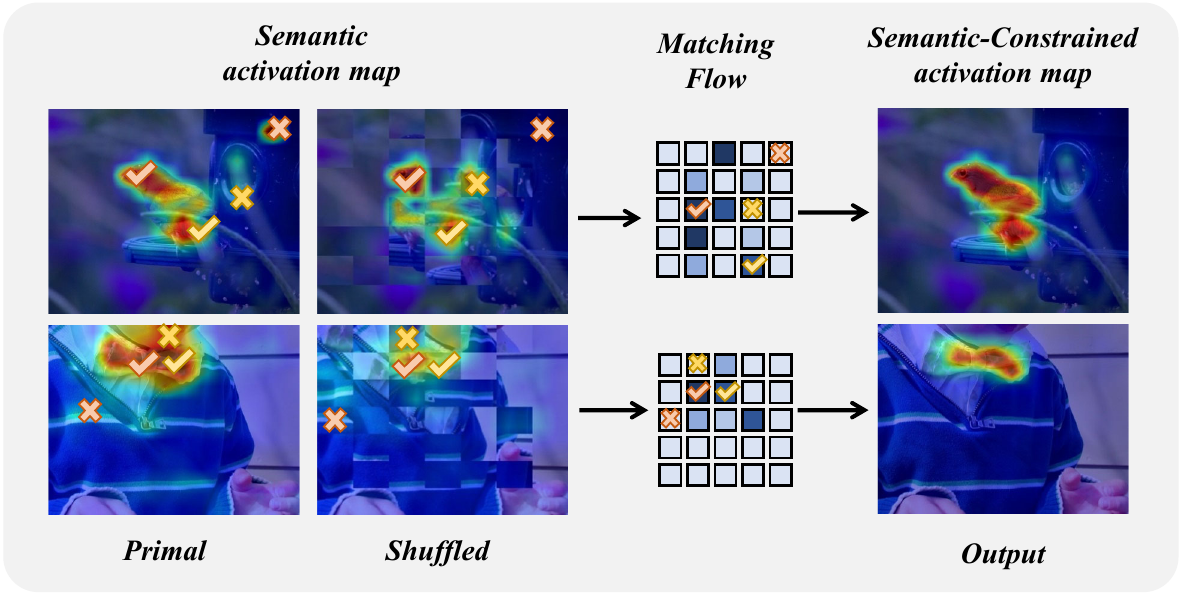}
    \caption{\textbf{Close-up visualization of the matching module.} The effect of more semantic restrictions is shown, where \Checkmark and \XSolidBrush denote the matched and unmatched regions in the primal-shuffled pair, respectively, and the same color indicates that it belongs to the same match. As expressed in the matching flow matrix, unmatched parts have smaller while matched parts have higher values of the corresponding matrix elements. According to the visualization, matched and unmatched regions tend to be foreground and background, respectively. Accordingly, the background activation (i.e., divergent activation) can be removed after training and calibrating the network. (Best viewed in color)}
    \label{Fig.main_4}
\end{figure}

Given the input primal-shuffled feature pair, the semantic-constraint matching module seeks to estimate the dense matching flow matrix,
which contains co-activation semantic knowledge. Since the primal-shuffled semantic activation map generally has high activation values in object region, the flow contains more co-object details than background. 
Therefore, we utilize regularization during training to let the network implicitly learn how to constrain divergent activations,
more visualization and analysis are shown in Figure \ref{Fig.main_4}.

\textbf{Overview.}
As mentioned in Sec.\ref{sec2.2}, we consider semantic-constraint matching as an optimal transport problem. First, analogous class activation maps $F'_p$, $F'_h$ are concatenated and unified to go through an across-transformer module for feature synthesis. This across-transformer module extracts the co-attention features $O\in\mathbb{R}^{D\times 2hw}$ with a global view using all patch tokens from the primal-shuffled pairs as input. After a straightforward split and reshape, we obtain the integrated features $O_p, O_h\in\mathbb{R}^{D\times h\times w}$ of the primal-shuffled pair. The matching flow is then obtained by modelling the optimal transport. Eventually, the matching flow is coupled with the integrated features to acquire the final optimized features.

\textbf{Preliminary.}
 The purpose of optimal transport is to calculate the minimum transportation cost between the source distribution $\zeta_s$ and target distribution $\zeta_t$. Consider these two as discrete probability distributions defined in probability space $X_s,X_t\in \Omega$, respectively, which can be written as:
\begin{equation}\label{eq3}
    \zeta_s=\sum_k^{N_s}\delta(x_s^k)\cdot \eta_s^k,~~~~~~ \zeta_t=\sum_k^{N_t}\delta(x_t^k)\cdot \eta_t^k
\end{equation}
where $N_s$ and $N_t$ represent the sample numbers, $\delta(.)$ means Dirac function and $\eta_s^k,\eta_t^k$ are the k-th sample's probability that belongs to the probability simplex space ($\sum_k^{N_s}\eta_s^k=\sum_k^{N_t}\eta_t^k=1$).
Typically, the Kantorovich function~\cite{kantorovich2006translocation} addresses optimal transport by computing a probability space $\pi(x_s,x_t)$ with a meaningful cost function $c(x_s,x_t)$:
\begin{equation}\label{eq4}
    \hat{\pi}=\mathop{\arg\min}_{\pi}\int_{X_s}\int_{X_t}\pi(x_s,x_t)\cdot c(x_s,x_t)d{x_s}d{x_t}
\end{equation}
where $\pi$ is the joint probability density, and $\zeta_s=\int_{X_s}\pi(x_s,x_t)d{x_s}$, $\zeta_t=\int_{X_t}\pi(x_s,x_t)d{x_t}$. By combining the cost function values of each sample, we can generate a cost matrix $C$, where $C_{ij}$ denotes the distance between $x_s^i$ and $x_t^j$. The optimal transport problem is:
\begin{equation}\label{eq5}
    \hat{T}=\mathop{\arg\min}_{T}\sum_{i,j}^{N_s,N_t}T_{ij}C_{ij}
\end{equation}

where $\hat{T}\in\mathbb{R}^{N_s\times N_t}$ designates the optimal transport matrix. $T\in\mathbb{R}^{N_s\times N_t}$ and $T_{ij}$ denotes the optimal transmission with minimum cost for moving from $x_s^i$ to $x_t^j$ ($TI_{N_t}=\zeta_s$, $TI_{N_s}=\zeta_t$).

\textbf{Optimal transport.}
In this work, we introduce a pairwise matching probability matrix $T\in\mathbb{R}^{hw\times hw}$ and build a cost matrix $\Gamma\in\mathbb{R}^{hw\times hw}$ from the source primal image to the target shuffled image. Our goal is to minimize the matching difference to gain a global optimal matching flow matrix $\hat{T}$. We borrow semantic activation maps $M_p$ and $M_h$ to generate empirical distributions $\zeta_s$ and $\zeta_t$, whose values indicate the importance of each feature vector in the feature map. Thus, we have a standard optimal transport problem as in Eq.\ref{eq5}.

The cost matrix represents the cost of transferring the source domain to the target domain, and a common approach to images is to compute the cosine similarity between image features. Here, we consider $O_p$ and $O_h$ as the final output features of the primal-shuffled pair, the cost matrix is computed as:
\begin{equation}\label{eq6}
    \Gamma=1-\frac{O_p\cdot O_h^\top}{\|O_p\|\|O_h\|}
\end{equation}
where $\Gamma\in\mathbb{R}^{hw\times hw}$ denotes the matching difference between the corresponding position in source primal feature map and target shuffled feature map.

Due to the similarity between semantic activation maps and CAM, we borrow the idea of hierarchical staircase function from~\cite{liu2020semantic}. In brief, with the staircase function, we cluster the pixels of the semantic activation graph into four groups: \textit{specific, general, uncertain and background} as:
\begin{equation}\label{eq7}
    (A_p, A_h)=\sum_i^4\beta_i(\psi(M_p>\alpha_i),\psi(M_h>\alpha_i))
\end{equation}

where $\alpha_i$ and $\beta_i$ is the i-th threshold and weight of staircase function. We define $\alpha=[0,0.4\mu,0.5\mu,0.6\mu]$ and $\beta=[0,0.8,0.9,1.0]$, where $\mu$ is a hyper-parameter represent the stair scale. $\psi(.)$ is an indicator function that outputs 1 only if the conditions in parentheses are met, and 0 otherwise. Accordingly, $\zeta_s$ and $\zeta_t$ can be calculated and flattened to vector samples as:
\begin{equation}\label{eq8}
    (\zeta_s,\zeta_t)=(A_p/\sum A_p, A_h/\sum A_h)
\end{equation}

The conventional solution for Eq.\ref{eq5} is NP-hard and inefficient. We perform entropy regularization following~\cite{cuturi2013sinkhorn}, thus Eq.\ref{eq5} can be reformulated as:
\begin{equation}\label{eq9}
\begin{split}
    \hat{T}=\mathop{\arg\min}_{T}\sum_{i,j}T_{ij}\Gamma_{ij}+\varepsilon\Phi(T)\\
    s.t.~\Phi(T)=\sum T(\log T-1)
\end{split}
\end{equation}
where $\Phi(T)$ is the entropy regularization and $\varepsilon$ is the regularization weight.The Sinkhorn-Knopp algorithm~\cite{sinkhorn1967diagonal,su2023occlusion} can easily solve convex optimization problems like Eq.\ref{eq9}.
 Since it contains solely matrix operations, it can be efficiently solved with hardware acceleration.

\begin{table}[!t]
    \centering
    \begin{tabular}{cc|ccc}
        \toprule
        \textit{Local shuffle} & \textit{SC Matching} & $Top1\ Loc.$ & $Top5\ Loc.$ & $GT\ Loc.$ \\
        \midrule
         &  & 70.0 & 83.5 & 87.8 \\
        \Checkmark &  & 71.2 & 84.5 & 88.5 \\
         & \Checkmark & 70.6 & 84.0 & 88.3 \\
         \rowcolor{gray!20}
        \Checkmark & \Checkmark & \textbf{73.0} & \textbf{85.3} & \textbf{89.1} \\
        \bottomrule
    \end{tabular}
    \caption{Exploration of different component combinations of network for training on the CUB-200-2011 dataset.}
    \label{tab:1}
\end{table}

\begin{table}[!t]
    \centering
    \setlength{\tabcolsep}{3.5mm}
    \begin{tabular}{cccc}
        \toprule
        \textit{Methods} & $Top1\ Loc.$ & $Top5\ Loc.$ & $GT\ Loc.$ \\
        \midrule
        EMD\cite{zhang2020deepemd} & 66.4 & 83.4 & \textbf{89.9} \\
        GW\cite{zhang2021deepacg} & 69.2 & 83.9 & 88.6  \\ 
        \rowcolor{gray!20}
        SC matching\textbf{(ours)} & \textbf{73.0} & \textbf{85.3} & 89.1 \\
        \bottomrule
    \end{tabular}
    \caption{Exploration of different mechanisms for matching on the CUB-200-2011 dataset.}
    \label{tab:2}
\end{table}

\textbf{Refine features.}
Conclusively, matching flow $\hat{T}$ is normalized column-wise and row-wise to obtain the attention maps across $O_p$($O_h$) for each position in $O_h$($O_p$), respectively. As the last optimization feature, we compute the attention synthesis:
\begin{equation}\label{eq13}
\begin{split}
    T_p=O_p\cdot softmax(\hat{T}), T_h=O_h\cdot softmax(\hat{T}^\top)
\end{split}
\end{equation}
where $T_p$ and $T_h$ are reshaped into $\mathbb{R}^{c\times h\times w}$. Common semantic knowledge and co-activated objects are preserved through optimized features. As a result, the network can comprehend object patterns more accurately and clearly after classification training.

\subsection{Network Training and Inference}~\label{sec3.4}
In this section we present two essential loss functions for end-to-end training of SCMN and some inference details. 

During training, as image-level classification label is the only human-annotated supervision that can be used here, we first define the basic classification loss function as:
\begin{equation}\label{eq10}
    \mathcal{L}_{cls}=\mathcal{L}_{ce}(y,\hat{y}_F) + \mathcal{L}_{ce}(y,\hat{y}_T)
\end{equation}
where $\mathcal{L}_{ce}$ is the cross-entropy loss used to update the gradient propagation for semantic information, $y$ is the ground-truth class and $\hat{y}=(\hat{y}_p,\hat{y}_h)$ indicate the prediction. And the subscripts $F$ and $T$ of $\hat{y}$ denote the predictions we obtained after GAP on the basis of $(F'_p, F'_h)$ and $(T_p, T_h)$, respectively. The first item of $\mathcal{L}_{cls}$ is the loss function suggested in TS-CAM~\cite{gao2021ts}, which ensures the network's underlying performance will remain steady. The second one, based on refined features, enables the network to link semantic labels and foreground regions more effectively. Furthermore, it fully utilizes the context of the training data to direct the network to comprehend the relationship between object parts and background, calibrating the Transformer to constrain divergent activation based on the resolution of partial activation issue.

Besides, the matching flow matrix $\hat{T}$ refined from the semantic-constraint matching module can be viewed as a domain migration matrix between the source and target domains.
To prevent the output from distorting and falling into local minimum, we introduce the equivariant regularization loss with reference to SEAM~\cite{wang2020self}:
\begin{equation}\label{eq11}
    \mathcal{L}_{er}={\|T_p-M_h\|}_1+{\|T_h-M_p\|}_1
\end{equation}
where $\|.\|_1$ means the $l1$ loss. Consequently, we can obtain implicit covariant constraints that allow the network to learn implicitly how to constrain the divergent activation.

By jointly optimizing the classification (Eq.\ref{eq10}) and the equivariant regularization (Eq.\ref{eq11}), the final training loss is defined as:
\begin{equation}\label{eq12}
    \mathcal{L}_{total}=\mathcal{L}_{cls}+\lambda\mathcal{L}_{er}
\end{equation}
where $\lambda$ is a hyper-parameter to balance the optimization.

In inference, we follow CAM~\cite{zhou2016learning} using a threshold approach to generate localization boxes. Note that to prevent the chaos of results from the randomness of the local patch shuffle strategy, we use the input itself to capture the final semantic-constrained activation map. Specifically, we directly input the primal image into the calibrated Transformer network to obtain the semantic-constrained activation map. Additionally, we discovered through testing that, in some cases, adding the across-transformer mentioned in the semantic restrictions module can improve outcomes. Sec.\ref{sec.4} describes the specific experimental results in detail.

\section{Experiments}
\label{sec.4}

\begin{table}[!t]
    \centering
    \setlength{\tabcolsep}{3.5mm}
    \begin{tabular}{cccc}
        \toprule
        \textit{Patch shuffle} & $Top1\ Loc.$ & $Top5\ Loc.$ & $GT\ Loc.$ \\
        \midrule
        Global & 67.1 & 80.5 & 84.5 \\
        baseline & 70.0 & 83.5 & 87.8 \\
        \rowcolor{gray!20}
        Local\textbf{(ours)} & \textbf{73.0} & \textbf{85.3} & \textbf{89.1} \\
        \bottomrule
    \end{tabular}
    \vspace{5pt}
    \caption{Exploration of global/local patch shuffle strategy on the CUB-200-2011 dataset.}
    \label{tab:3}
\end{table}

\begin{table}[!t]
    \centering
    \setlength{\tabcolsep}{3.5mm}
    \begin{tabular}{cccc}
        \toprule
        Stair Scale $\mu$ & $Top1\ Loc.$ & $Top5\ Loc.$ & $GT\ Loc.$ \\
        \midrule
        0.5 & 72.4 & 84.8 & 88.8 \\
        0.7 & 72.5 & 85.1 & 89.0 \\
        \rowcolor{gray!20}
        1.0 & \textbf{73.0} & \textbf{85.3} & \textbf{89.1} \\
        \bottomrule
    \end{tabular}
    \caption{Exploration of different scale in staircase function on the CUB-200-2011 dataset.}
    \label{tab:4}
\end{table}

\subsection{Experimental Settings}~\label{sec.4.1}
\textbf{Datasets.}
Consistent with other methods, SCMN is evaluated on two challenging benchmarks, including CUB-200-2011~\cite{wah2011caltech} and ILSVRC2012~\cite{russakovsky2015imagenet}, with only category labels for training. CUB-200-2011 is a fine-grained dataset containing 200 varieties of birds, which is divided into the training set of 5,994 images and the testing set of 5,794 images. ILSVRC2012 is a larger scale classification dataset with 1000 categories, consisting of 1,281,197 training images and 50,000 validation images.

\textbf{Evaluation Metrics.}
Weakly supervised object localization focus on the localization performance of the network for a given class. Following the baseline methods~\cite{zhou2016learning,pan2021unveiling,gao2021ts}, we apply three kinds of metrics to evaluate bounding box prediction: Top-1/Top-5 localization accuracy (\textit{Top-1/Top-5 Loc.}) and localization accuracy with ground-truth class (\textit{GT Loc.}). For the former, a prediction is positive when it concurrently satisfies the following two conditions: the predicted classification labels match the ground-truth categories; the predicted bounding boxes have over 50\% IoU with at least one of the ground-truth boxes. Moreover, the latter indicates that only the accuracy of the final generated bounding box is considered while ignoring the classification results.

\begin{table}[!t]
    \centering
    \setlength{\tabcolsep}{3.5mm}
    \begin{tabular}{c|c|c|c|c}
        \toprule
        \multirow{2}{*}{Methods} & \multirow{2}{*}{Backbone} & \multicolumn{3}{c}{$Loc\ Acc.$} \\
        ~ & ~ & \textit{Top1.} & \textit{Top5.} & \textit{GT-K.} \\
        \midrule
        CAM~\cite{zhou2016learning} & VGG16 & 41.1 & 50.7 & 55.1 \\
        ACoL~\cite{zhang2018adversarial} & VGG16 & 45.9 & 56.5 & 59.3 \\
        SPG~\cite{zhang2018self} & VGG16 & 48.9 & 57.2 & 58.9 \\
        DANet~\cite{xue2019danet} & VGG16 & 52.5 & 62.0 & 67.7 \\
        ADL~\cite{choe2019attention} & VGG16 & 52.4 & - & 75.4  \\
        MEIL~\cite{mai2020erasing} & VGG16 & 57.5 & - & 73.8 \\
        DGL~\cite{tan2020dual} & VGG16 &  56.1  &  68.5  &  74.6  \\
        SPA~\cite{pan2021unveiling} & VGG16 &  60.2  &  72.5  &  77.2  \\
        SLT-Net~\cite{guo2021strengthen} & VGG16 & 67.8 &  -  & 87.6\\
        CI-CAM~\cite{shao2021improving} & VGG16 & 58.4 & 70.5  & 75.7  \\
        \midrule
        CAM~\cite{zhou2016learning} & GoogLeNet & 41.1 & 50.7 & 55.1 \\
        SPG~\cite{zhang2018self} & GoogLeNet &46.7 & 57.2 & - \\
        DANet~\cite{xue2019danet} & InceptionV3 &49.5 & 60.5 & 67.0 \\
        ADL~\cite{choe2019attention} & InceptionV3 &53.0  &-  &- \\
        PSOL~\cite{zhang2020rethinking} & InceptionV3 &65.5 & - & - \\
        DGL~\cite{tan2020dual} & InceptionV3 & 50.5  & 62.2  & 67.6  \\
        SPA~\cite{pan2021unveiling} & InceptionV3 &53.6 & 66.5 & 72.1 \\
        SLT-Net~\cite{guo2021strengthen} & InceptionV3 &66.1 & - & 86.5 \\
        \midrule
        TS-CAM~\cite{gao2021ts} & Deit-S & 71.3 & 83.8 & 87.7 \\
        SCMN\textbf{(ours)} & Deit-S & \color{blue}\textbf{73.0} & \color{blue}\textbf{85.3} & \color{blue}\textbf{89.1} \\
        SCMN$^\dagger$\textbf{(ours)} & Deit-S & \color{red}\textbf{77.3} & \color{red}\textbf{89.4} & \color{red}\textbf{92.9} \\
        \bottomrule
    \end{tabular}
    \caption{Comparison of localization accuracy with the state-of-the-arts on the CUB-200-2011 test set. $^\dagger$indicates that across-transformer is added. The best two results on each dataset are shown in {\color{red}red} and {\color{blue}blue}.}
    \label{tab:5}
\end{table}

\textbf{Implementation Details.}
For fair comparisons, we adopt the same transformer backbone~\cite{touvron2021training} as TS-CAM~\cite{gao2021ts}.
Specifically, we release the top MLP layer and insert one convolution layer with kernel size 3$\times$3, stride 1, padding 1, and 200 (ILSVRC of 1000) dimension output following Gao's approach~\cite{gao2021ts}. A global average pooling (GAP) layer is added at the top. During training, each input image is reshaped to $256\times256$ and randomly cropped to $224\times224$. We then use AdamW with $\epsilon=1e-8$, $\beta_1=0.9$, $\beta_2=0.99$ and weight decay of 5e-4. On CUB-200-2011, we train 60 epochs with $\lambda=0.5$, learning rate of 1e-5, batch size of 64. On ILSVRC2012, we train 20 epochs with $\lambda=0.1$, learning rate of 1e-5 and batch size of 64.

\subsection{Ablation Study}
In this section, we implement several ablation studies to verify the efficacy of SCMN. 
Note that semantic-constraint matching module is rename as \textit{SC Matching} for simplicity. All experiments are conducted on the CUB-200-2011 dataset.

\textbf{The effect of each component.}
As reported in Table \ref{tab:1}, we investigate the accuracy of different network configurations on the CUB-200-2011 test set. Note that we replace it with a $1\times1$ convolution operation when there is no semantic-constraint matching module. Furthermore, when without local patch shuffle strategy, we use the primal image and its own for matching. It can be seen that the local patch shuffle strategy increases the Top1/Top5/Gt loc by 1.7\%/1.2\%/0.8\% above the existing baseline, reflecting the improved robustness of the network. At the same time, matching only the image itself can also capture the highlighted part of the object, which is reflected in the improvement of every metrics. The best localization accuracy can be achieved when both of them are implemented.

\textbf{The effect of different matching mechanisms.}
We reveal the advantages by comparing semantic-constraint matching with other classical matching techniques. Following~\cite{zhang2020deepemd, zhang2021deepacg}, we adopt the Earth Mover’s Distance (EMD) and Gromov-Wasserstein (GW) distance to build dense correlation volumes for all pairs of image. As shown in Table \ref{tab:2}, EMD improves localization performance significantly but is weak in classification, where GW is more vulnerable than our Semantic-constraint matching overall.

\begin{table}[!t]
    \centering
    \setlength{\tabcolsep}{3.5mm}
    \begin{tabular}{c|c|c|c|c}
        \toprule
        \multirow{2}{*}{Methods} & \multirow{2}{*}{Backbone} & \multicolumn{3}{c}{$Loc\ Acc.$} \\
        ~ & ~ & \textit{Top1.} & \textit{Top5.} & \textit{GT-K.} \\
        \midrule
        CAM~\cite{zhou2016learning} & VGG16 & 38.9 &  48.5 &  - \\
        ACoL~\cite{zhang2018adversarial} & VGG16 & 45.8 &  59.4 &  63.0 \\
        ADL~\cite{choe2019attention} & VGG16 & 44.9 &  - &  -  \\
        MEIL~\cite{mai2020erasing} & VGG16 & 46.8 &  - &  - \\
        DGL~\cite{tan2020dual} & VGG16 & 47.7  & 58.9 & 64.8  \\
        PSOL~\cite{zhang2020rethinking} & VGG16 &50.9 & 60.9 & 64.0\\
        SPA~\cite{pan2021unveiling} & VGG16 & 49.6 &  61.3 &  65.1  \\
        SLT-Net~\cite{guo2021strengthen} & VGG16 &51.2 & 62.4 & 67.2\\
        CI-CAM~\cite{shao2021improving} & VGG16 & 48.7 & 58.8  & 62.4  \\
        \midrule
        ACoL~\cite{zhang2018adversarial} & GoogLeNet & 46.7 &  57.4 &  -\\
        DANet~\cite{xue2019danet} & GoogLeNet & 47.5 &  58.3 &  -\\
        CAM~\cite{zhou2016learning} & InceptionV3 &46.3 & 58.2 & 62.7  \\
        SPG~\cite{zhang2018self} & InceptionV3 &48.6 & 60.0 & 64.7 \\
        MEIL~\cite{mai2020erasing} & InceptionV3 &49.5 & - & -  \\
        GC-Net~\cite{lu2020geometry} & InceptionV3 &49.1 & 58.1 & -  \\
        DGL~\cite{tan2020dual} & InceptionV3 & 52.2 & 63.4 & \color{blue}68.1  \\
        PSOL~\cite{zhang2020rethinking} & InceptionV3 &\color{red}54.8 & 63.3 & 65.2 \\
        SPA~\cite{pan2021unveiling} & InceptionV3 & 52.8 &  64.3 & \color{red}68.4\\
        \midrule
        TS-CAM~\cite{gao2021ts} & Deit-S & 53.4 & 64.3 & 67.6 \\
        SCMN\textbf{(ours)} & Deit-S & \color{red}\textbf{54.8} & \color{red}\textbf{65.2} & \color{red}\textbf{68.4} \\
        SCMN$^\dagger$\textbf{(ours)} & Deit-S & \color{blue}\textbf{54.3} & \color{blue}\textbf{64.6} & \textbf{67.6} \\
        \bottomrule
    \end{tabular}
    \caption{Comparison of localization accuracy with the state-of-the-arts on the ILSVRC2012 validation set. $^\dagger$indicates that across-transformer is added. The best two results on each dataset are shown in {\color{red}red} and {\color{blue}blue}.}
    \label{tab:6}
\end{table}

\textbf{The effect of local/global shuffle.}
As we mentioned in Sec.\ref{sec3.2}, the global shuffling approach can be detrimental to the network performance, while our proposed local patch shuffle mechanism can better constrain the spreading activation. The comparison of quantitative metrics is shown in Table \ref{tab:3}, we observed that local shuffle strategy outperforms the baseline (the first setting in Table \ref{tab:1}) by 4.3\% when global shuffle underperforms 4.1\%.

\textbf{The effect of threshold in Semantic-constraint matching module.}
In our proposed semantic-constraint matching module, we use a staircase function for the semantic activation map to group it into four grades, where $\mu$ is an adjustable hyper-parameter. In Table \ref{tab:4}, we analyze the effect of different values of $\mu$ on the final results, while $\mu=1.0$ we reach the highest performance.

\subsection{Evaluation}

\begin{figure*}[!t]
  \centering
  \includegraphics[width=\textwidth]{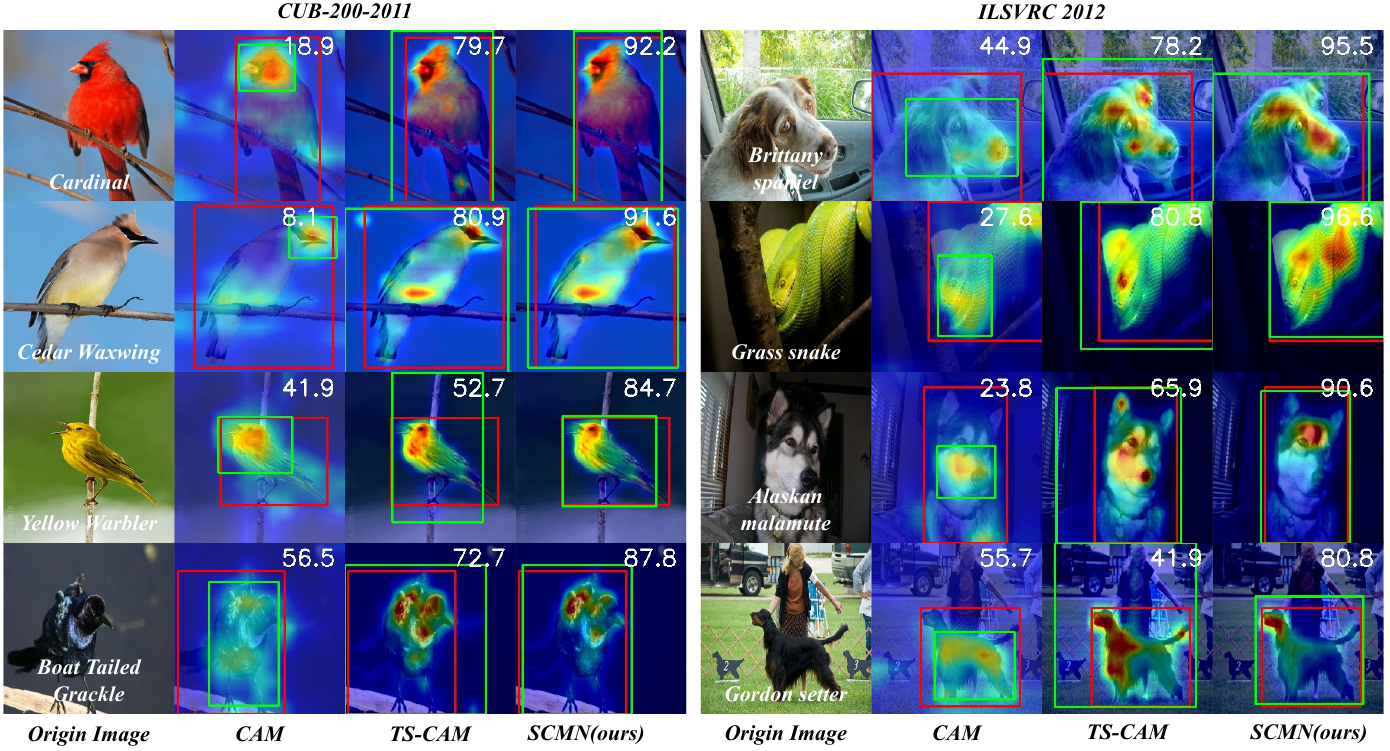}
  \caption{Qualitative comparison on the localization results of the CUB-200-2011 and ILSVRC2012 datasets. Notice that each method's predictions are in {\color{green}green}, the ground truth are in {\color{red}red}, and their IOU values are in white text. Our SCMN achieve better results while CAM tends to focus on the discrimination part, and TS-CAM activate more background regions.}
  \label{Fig:main_5}
\end{figure*}

\textbf{Quantitative result.}
We first quantitatively compare the proposed SCMN with the state-of-the-arts separately on the CUB-200-2011 and ILSVRC2012 datasets, as displayed in Table~\ref{tab:5} and Table~\ref{tab:6}. On CUB-200-2011 test set, we attended that our method significantly outperforms the CNN-based method and is superior to TS-CAM in all three metrics by 2.4\%/1.8\%/1.6\% when using the same Deit-S backbone. 
In addition, SCMN with Deit-S backbone have achieved more comprehensive and competitive results on ImageNet than advanced methods. It improved by 2.6\%/1.4\%/1.2\% compared with TS-CAM. As mentioned in Sec.\ref{sec3.4}, we found a significant improvement in the results of the CUB-200-2011 dataset after adding across-transformer, with a 8.4\%/6.7\%/5.9\% improvement compared to TS-CAM. However, in the ILSVRC2012 dataset, the effect is reduced instead. Since the CUB-200-2011 dataset is all birds with slight interclass disparity, the across-transformer can learn the association between the foreground and background of different classes more efficiently. In contrast, the ILSVRC2012 dataset contains numerous categories, making it harder for across-transformers to learn a good representation, thus reducing the performance.

\textbf{Qualitative result.}
For qualitative result, Figure \ref{Fig:main_5} visualizes the localization outcomes of various pipelines of different classes in the CUB-200-2011 and ILSVRC2012 datasets. We followed TS-CAM using the same threshold of 0.1 to generate the results.
From the visualization, it is clear that our strategy no longer focuses solely on the object's local part but covers the entity's whole body region more completely compared with CAM. Furthermore, analogized with TS-CAM, which is also based on Transformer, our approach effectively prevents invalid background coverage by suppressing its implicit presence of divergent activation problem. For example, with \textit{Yellow Warbler} and \textit{Gordon setter}, CAM only locates the most prominent parts like head and body. In contrast, TS-CAM additionally discovers background areas such as tree branches, people, and grass. On the other hand, our method is very close to the ground truth with a high IOU value.

\subsection{Further Downstream Application}
In this part, we investigate SCMN's adaptability to other tasks. For example, weakly supervised salient object detection (WSSOD) involves locating the most visually prominent or salient objects with less precise or incomplete labels, such as image-level tags or points indicating the presence of an object. Many of these works~\cite{wang2017learning, li2018weakly, zeng2019multi, piao2021mfnet} use category labels for salient detection because of the availability of large datasets covering multiple categories, such as ImageNet. Therefore, we can apply SCMN to WSSOD based on this.

\begin{table*}[!t]
    \centering
    \resizebox{\textwidth}{!}{
    \begin{tabular}{c|cccc|cccc|cccc|cccc}
        \toprule
        \multirow{2}{*}{Methods} & \multicolumn{4}{c|}{DUTS} & \multicolumn{4}{c|}{DUT-OMRON} & \multicolumn{4}{c|}{ECSSD} & \multicolumn{4}{c}{HKU-IS} \\
        ~ & \textit{$F_\beta$} & \textit{$S_\alpha$} & \textit{$E_s$} & \textit{$M$} & \textit{$F_\beta$} & \textit{$S_\alpha$} & \textit{$E_s$} & \textit{$M$} & \textit{$F_\beta$} & \textit{$S_\alpha$} & \textit{$E_s$} & \textit{$M$} & \textit{$F_\beta$} & \textit{$S_\alpha$} & \textit{$E_s$} & \textit{$M$}  \\
        \midrule
        WSS~\cite{wang2017learning} & 0.654 & 0.748 & 0.795 & 0.100 & 0.603 & 0.725 & 0.768 & 0.109 & 0.823 & 0.811 & 0.869 & 0.104 & 0.821 & 0.822 & 0.896 & 0.079 \\
        ASMO~\cite{li2018weakly} & 0.614 & 0.697 & 0.772 & 0.116 & 0.622 & 0.752 & 0.776 & 0.101 & 0.797 & 0.802 & 0.853 & 0.110 & - & - & - & -  \\
        MSW~\cite{zeng2019multi} & 0.684 & 0.759 & 0.814 & 0.091 & 0.609 & 0.756 & 0.763 & 0.109 & 0.840 & 0.827 & 0.884 & 0.096 & 0.814 & 0.818 & 0.895 & 0.084  \\
        MFNet~\cite{piao2021mfnet} & \color{blue}0.710 & \color{blue}0.775 & \color{blue}0.839 & \color{blue}0.076 & 0.646 & 0.742 & 0.803 & 0.087 & 0.854 & \color{blue}0.834 & 0.885 & \color{blue}0.084 & 0.851 & \color{blue}0.846 & \color{blue}0.921 & \color{blue}0.059  \\
        \midrule
        SCMN-MF & \textbf{0.705} & \textbf{0.768} & \textbf{0.836} & \textbf{0.080}   & \color{blue}\textbf{0.661} & \color{blue}\textbf{0.755} & \color{blue}\textbf{0.815} & \color{blue}\textbf{0.082}   & \color{blue}\textbf{0.856} & \textbf{0.832} & \color{blue}\textbf{0.890} & \textbf{0.086}   & \color{blue}\textbf{0.854} & \textbf{0.839} & \textbf{0.919} & \textbf{0.062}  \\
        SCMN-MF$^\dagger$ & \color{red}\textbf{0.736} & \color{red}\textbf{0.776} & \color{red}\textbf{0.847} & \color{red}\textbf{0.074}   & \color{red}\textbf{0.691} & \color{red}\textbf{0.761} & \color{red}\textbf{0.825} & \color{red}\textbf{0.077}   & \color{red}\textbf{0.864} & \color{red}\textbf{0.840} & \color{red}\textbf{0.894} & \color{red}\textbf{0.071}   & \color{red}\textbf{0.868} & \color{red}\textbf{0.854} & \color{red}\textbf{0.925} & \color{red}\textbf{0.047}  \\
        \bottomrule
    \end{tabular}}
    \caption{Comparison of salient object detection metrics over four datasets with the state-of-the-arts.$^\dagger$ indicates that the results were further processed by CRF. The best two results on each dataset are shown in {\color{red}red} and {\color{blue}blue}.}
    \label{tab:7}
\end{table*}

\textbf{Approach.}
Similar to WSOL, WSSOD necessitates the creation of pseudo labels via CAM, so the pseudo-label is also inevitably affected by feature deviation, exhibiting problems such as partial activation. 
Thus, SCMN can provide finer pseudo labels. 
To get salient detection results, we combine SCMN with the outstanding pioneer work MFNet~\cite{piao2021mfnet}. We build pseudo labels based on the semantic-constrained activation graph produced by SCMN. 
The MFNet saliency network is trained using the obtained pseudo labels to produce the ultimate salient detection results.

\textbf{Evaluation.}
In the interests of equity, we follow MFNet by adopting ImageNet~\cite{russakovsky2015imagenet} and DUTS~\cite{wang2017learning} datasets as our training sets of SCMN and saliency network, respectively. 
We test on
the most commonly used datasets, DUTS~\cite{wang2017learning}, DUT-OMRON~\cite{yang2013saliency}, ECSSD~\cite{yan2013hierarchical}, and HKU-IS~\cite{li2015visual}.
In keeping with other research, we employ four measurements, including F-measure~\cite{achanta2009frequency}, S-measure~\cite{fan2017structure}, E-measure~\cite{fan2018enhanced} and mean absolute error (MAE), to assess the effect thoroughly. 

\begin{figure}[!t]
    \centering
    \includegraphics[width=\columnwidth]{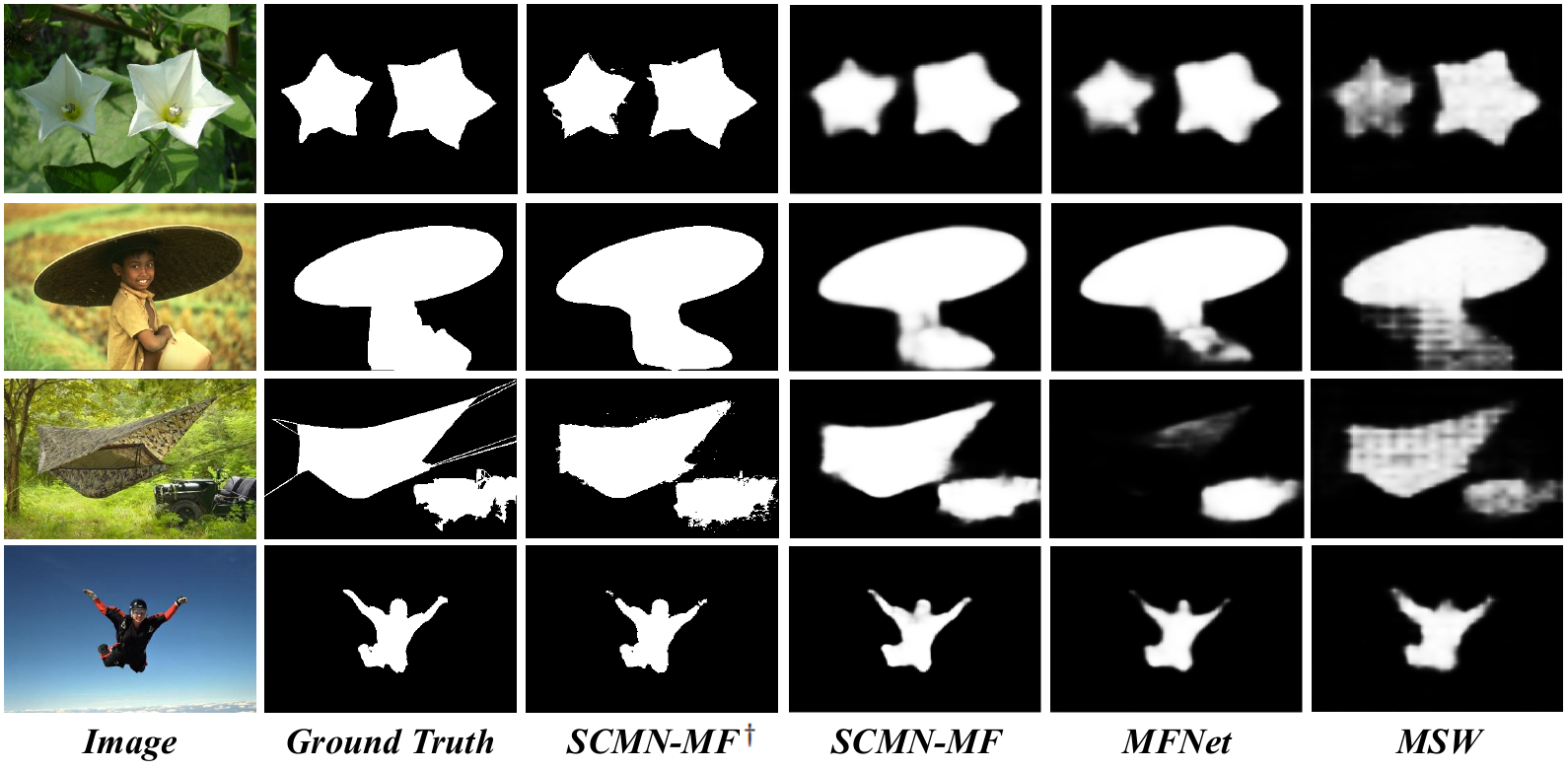}
    \caption{Qualitative comparison on the salient object detection results in some challenging scenes.}
    \label{Fig.main_6}
\end{figure}

Table \ref{tab:7} compares the quantitative results across the four datasets. Without further post-processing of the results, our effects are improved on DUT-OMRON, ECSSD, and HKU-IS, compared to the baseline MFNet, and are far superior to the other methods. Furthermore, our strategy achieves state-of-the-art when the post-processing technique is included. Besides, we show some examples in Figure \ref{Fig.main_6}. Our approach can find more comprehensive and clearly defined targets compared to other approaches. 


\section{Conclusion}
In this paper, we propose a novel Semantic-Constraint Matching Transformer Network for weakly supervised object localization. 
By exploring the underlying reasons for traditional methods, we first discard the classical CNN in favor of the Transformer architecture with long-range feature dependencies. Then, we propose a local patch shuffle strategy tailored for visual transformers and silence the divergent activation problem embedded in the transformer by a semantic-constraint matching module.
Extensive evaluations exhibit the efficacy of our system and we can achieve new state-of-the-art results. We hope these conclusions will encourage more research on weakly supervised object localization.


\section*{Acknowledgement}
This work was supported by National Natural Science Foundation of China (NSFC) 61876208 and 62272172, Tip-top Scientific and Technical Innovative Youth Talents of Guangdong Special Support Program (2019TQ05X200) and 2022 Tencent Wechat Rhino-Bird Focused Research Program Research (Tencent WeChat RBFR2022008).

\bibliographystyle{IEEEtran}
\bibliography{ref}

\end{document}